\documentclass[10pt,conference]{IEEEtran}
\usepackage[utf8]{inputenc}

\usepackage{graphicx}
\usepackage{multicol}
\usepackage[dvipsnames]{xcolor}
\usepackage{url, hyperref}
\usepackage[space]{cite}
\usepackage{comment}

\newcommand{\CV}{\texttt{code2vec}}
\newcommand{\GN}{\texttt{GGNN}}

\newcommand{\Part}[1]{\textbf{#1}}

\title{Testing Neural Program Analyzers}
\author{
    \textbf{Md Rafiqul Islam Rabin}\\
    University of Houston\\
    mdrafiqulrabin@gmail.com
    \and
    \textbf{Ke Wang}\\
    Visa Research\\
    kewangad@gmail.com
    \and
    \textbf{Mohammad Amin Alipour}\\
    University of Houston\\
    amin.alipour@gmail.com
}

\begin{document}

\maketitle

\begin{abstract}

%Solving programming task is a significant area of research. 
Deep neural networks have been increasingly used in software engineering and program analysis tasks.
They usually take a program and make some predictions about it, e.g., bug prediction.
We call these models \emph{neural program analyzers}. 
%DL is considered as most promising techniques to solve the programming task. Several models have already been proposed to solve various programming task.
The reliability of neural programs can impact the reliability of the encompassing analyses. 

In this paper, we describe our ongoing efforts to develop effective techniques for testing neural programs.
We discuss the challenges involved in developing such tools and our future plans. 
In our preliminary experiment on a neural model recently proposed in the literature, 
% \CV{} \cite{Alon:Model:code2vec:CoRR:2018}, 
we found that the model is very brittle, and simple perturbations in the input can cause the model to make mistakes in its prediction.

%could result in different output on the transformed program compared to the output of the original program. This work signifies the benefits of program transformation in the generation of new test programs to verify the correctness of neural models.

\end{abstract}
\section{Introduction}
\label{sec:introduction}

The advances of deep neural models in software engineering and program analysis research have received significant attention in recent years.
Researchers have already proposed various neural models (e.g., 
Tree-LSTM \cite{Tai:Model:TreeLSTM:CoRR:2015}, 
Gemini \cite{Xu:Model:Gemini:CoRR:2017}, 
GGNN \cite{Allamanis:Model:GGNN:CoRR:2017}, 
Code Vectors \cite{Henkel:Model:CodeVectors:CoRR:2018}, 
code2vec \cite{Alon:Model:code2vec:CoRR:2018}, 
code2seq \cite{Alon:Model:code2seq:CoRR:2018}, 
DYPRO \cite{Wang:Model:DYPRO:CoRR:2019, Wang:Model:DynamicEmbedding:CoRR:2019}, LIGER \cite{Wang:Model:LIGER:arXiv:2019}, 
Import2Vec \cite{Theeten:Model:Import2Vec:MSR:2019}) 
to solve problems related to different program analysis or software engineering tasks.
Although each neural model has been evaluated by its authors, in practice, these neural models may be susceptible to untested test inputs.
Therefore, a set of testing approaches has already been proposed to trace the unexpected corner cases.
Recent neural model testing techniques include 
\cite{Tian:Transformation:DeepTest:ICSE:2018, Zhang:Transformation:DeepRoad:ASE:2018}
for models of autonomous systems, 
\cite{Rychalska:Adversarial:SQuAD:arXiv:2018, Ribeiro:Adversarial:SEARs:ACL:2018, Lei:Adversarial:Discrete:CoRR:2018}
for models of QA systems, and 
\cite{Wang:Transformation:COSET:arXiv:2019, Wang:Model:LIGER:arXiv:2019}
for models of embedding systems.
However, testing neural models that work on source code has received little attention from researchers except the exploration initiated by Wang et al. \cite{Wang:Transformation:COSET:arXiv:2019}.

Evaluating the robustness of neural models that process source code is of particular importance because their robustness would impact the correctness of the encompassing analyses that use them.
In this paper, we propose a transformation-based testing framework to test the correctness of state-of-the-art neural models running on the programming task.
The transformation mainly refers to the semantic changes in programs that result in similar programs.
The key insight of transformation is that the transformed programs are semantically equivalent to their original forms of programs but have different syntactic representations.
For example, one can replace a \texttt{switch} statement of a program with conditional \texttt{if-else} statements. The original program of the \texttt{switch} statement is semantically equivalent to the new program of \texttt{if-else} statements.
A set of transformations can be applied to a program to generate more semantically equivalent programs, and those new transformed programs can be evaluated on neural models to test the correctness of those models.

The main motivation to apply transformation is the fact that such transformations may cause the neural model to behave differently and mispredict the input.
We are conducting a small study to assess the applicability of transformations in the testing of neural models.
The preliminary results show that the transformations are very effective in finding irrelevant output in neural models.
We closely perceive that the semantic-preserving transformations can change the predicted output or the prediction accuracy of neural models compared to the original test programs.

\begin{comment}
    \textbf{Contributions}
    This paper makes the following main contributions:
    \begin{itemize}
        \item We discuss the problem of testing neural program. 
        
        \item We propose an approach based on program transformations.
        % (i.e. renaming variables, exchanging loop, swapping boolean, converting switch and permuting the order of statements).
    %    \item We present a tool to automatically generate transformed programs that can be used as effective test inputs.
    %    \item We define metamorphic relations.
        \item We provide a preliminary evaluation of \T{} on state-of-the-art neural model.
    %    \item Additionally, we plan to make our transformation framework along with the generated transformed programs available for public use.
    \end{itemize}
\end{comment}
\section{Motivating Example}
\label{sec:example}

\begin{figure} %[!ht]
    \centering
    \includegraphics[width=\columnwidth]{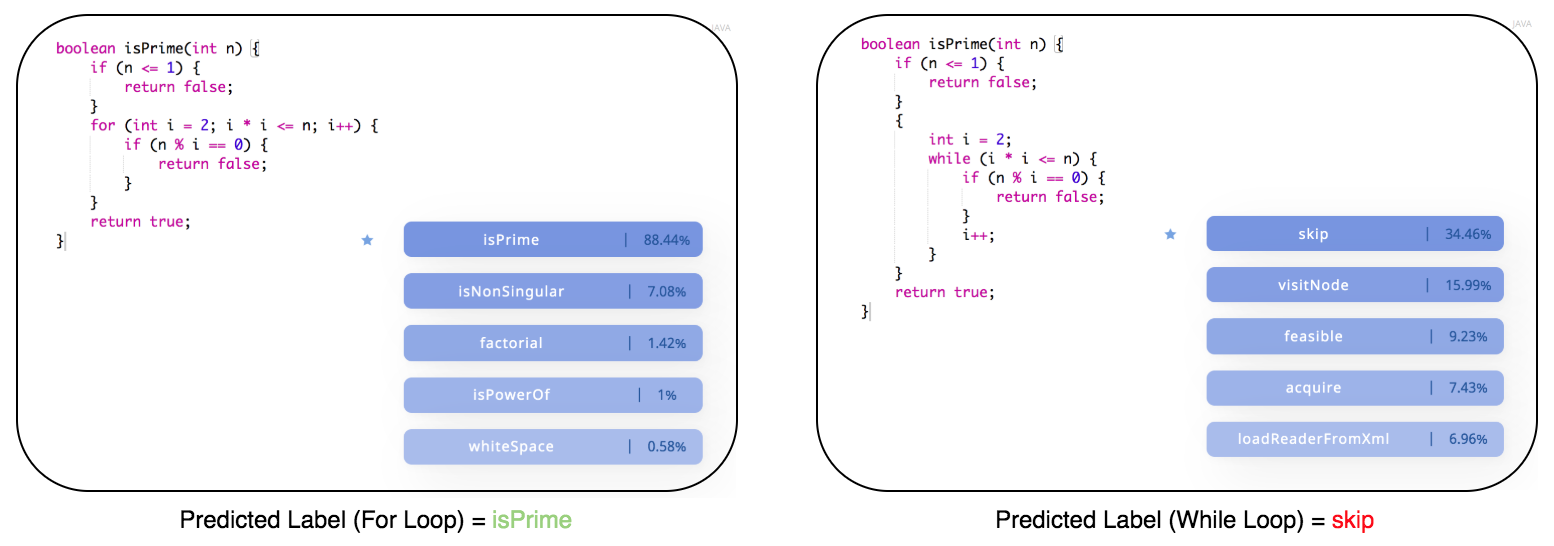}
    \caption{A failure in the \CV{} model revealed by a transformation.} 
    \label{fig:example}
\end{figure}

We use Figure~\ref{fig:example} as a motivating example to highlight the usefulness of our approach.
The code snippet shown in Figure~\ref{fig:example} is a simple Java method that demonstrates the prime functionality.
The functions check whether an integer is a prime number.
The only difference between these functions is that the implementation on the left uses a \texttt{for} loop, while the implementation on the right uses a \texttt{while} loop.

We instrument the prediction of the \CV{} model \cite{Link:code2vec:Demo} with these two equivalent functions.
The \CV{} takes a program and predicts its content.
\begin{comment}
    \CV{} correctly predicts the similarity for the implementation on the left, while it fails to produce correct results for the program on the right that is semantically equivalent.
    The right-program has been generated by applying the loop-exchanging transformation on left-program. 
    Both versions of the program are equivalent and doing the same functionality -- so we expect the identical behaviors from \CV{} model while evaluating these two versions of the program
\end{comment}
The result of the online demo \cite{Link:code2vec:Demo} reveals that the \CV{} model successfully predicts the program on the left as an ``isPrime" method, but cannot predict the program on the right as an ``isPrime" method. 
The model mistakenly predicts the program on the right as a ``skip" method, even though the ``isPrime" method is not included in the top-5 predictions made by the \CV{} model.

\begin{comment}
    Therefore, the transformation can be an effective approach to identify inconsistency in neural models.    
\end{comment}

\section{Proposed Methodology}
\label{sec:methodology}

In this section, we describe our efforts for testing neural programs. 
Currently, we are investigating semantic-preserving transformations that can potentially mislead a neural model of programs. 

\begin{comment}
    %Our approach is based on the syntactic transformation of programs to generate new semantically equivalent programs. 
    The approach takes programs as inputs and performs semantic-preserving transformation on those programs.
    The goal of \T{} is to generate synthetic test programs that make neural models to mistakenly produce wrong output.
\end{comment}

\begin{figure}%[!ht]
    \centering
    \includegraphics[width=\columnwidth]{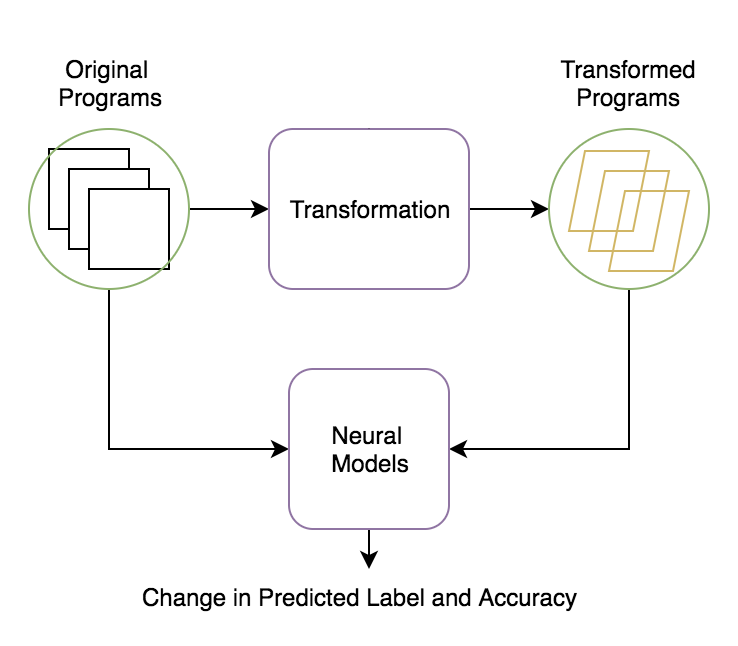}
    \caption{The workflow of our approach.}
    \label{fig:workflow}
\end{figure}

Figure~\ref{fig:workflow} depicts an overview of our approach for testing the neural models.
It can broadly be divided into two main steps: (1) Generating synthetic test programs using the semantic transformation of the programs in the original dataset, and (2) Comparing the predictions for the transformed programs with those for the original programs. 

%exhibit erroneous behaviors in neural models, and (3) analyze the results of the experiment.
%We describe these phases in the following subsections.

\Part{Semantic-Equivalent Program Transformations}
We have implemented multiple semantic program transformations to generate synthetic programs. 
Those semantic-preserving transformations include renaming variables, exchanging loops, swapping boolean values, converting switches and permuting the order of statements. 
In variable-renaming transformation, we rename all the occurrences of specific variables in program using an arbitrary name.
The boolean-swapping transformation refers to swapping \texttt{true} with \texttt{false} and vice versa, and we also neglect the condition so that the semantic is maintained.
In the same way, the loop-exchanging transformation means replacing a \texttt{while} loop with a \texttt{for} loop, and vice versa.
In switch-converting transformation, we replace the \texttt{switch} statements with the conditional \texttt{if-else} statements.
Finally, we include another transformation by permuting the order of statements without any semantic violations. 
All these transformations maintain semantic equivalence but generate different syntactic programs.
Thus far, we have not found any one transformation that works substantially better than others. 

%\subsection{Exhibit Erroneous Behaviours in Neural Models}
\Part{Test Oracle}
%We have designed an algorithm to automatically apply transformations on original programs to generate more semantically equivalent programs.
%After that, 
We evaluate both the original program and the transformed program in the neural model.
We mainly look at the predicted label and the prediction accuracy of the model for both original and transformed programs.
The neural model should behave similarly with both the original and the transformed program, which we define as a transformation-based metamorphic relation.
The main challenge in this phase is to define a measure for the \emph{similarity} of the predictions.
We are experimenting with a few ideas for this phase, for example, setting a threshold for the similarity of the predictions.

\Part{Challenges Ahead}
There are five main challenges that we are aiming to address in this project: (1) what types of transformation should be performed, (2) how to preserve the semantic equivalence during transformations, (3) where to apply those transformations, (4) how to control the transformation strategies, and (5) how to evaluate the transformed programs.

\begin{comment}
    
%Any violation of the metamorphic relation has been considered as erroneous behaviors in neural models.

\subsection{Analyze the Results of the Experiment}

    We have been experimenting on the \CV{} model \cite{Alon:Model:code2vec:CoRR:2018} with the online demo examples \cite{Link:code2vec:Demo} which contains 11 java files of a single method.
    At the beginning, we apply transformations on original dataset and generate the new transformed dataset. 
    Then we build the \CV{} repository \cite{Link:code2vec:Repository} and download an already-trained model for our initial study.
    After the model loads, we preprocess both the original dataset and the transformed dataset with JavaExtractor module that results in required c2v embedding for the model.
    We evaluate the model separately with the c2v embedding of the original and transformed dataset which writes the model's predictions and attention scores in a log file.
    We compare the predictions and attentions for identifying the transformed java file that makes the model to output different result than the original java file.
    The results show that \T{} is effective in finding several erroneous behaviors in the neural model of the programming task.

\end{comment}

\section{Our Plan}
\label{sec:future}

Thus far, we have applied five types of transformation. 
Those transformations are only capable of making basic changes in the syntactic representations of programs. 
However, our target is to devise more systematic transformations.
We are investigating the techniques and heuristics to suggest \emph{places} in programs to transform, and the \emph{types of transformation} that are most likely to cause the neural model to mispredict.  

Moreover, we have only evaluated our transformation on the \CV{} model \cite{Alon:Model:code2vec:CoRR:2018}, where the target task is to label the method name given a method body. 
We also plan to evaluate the transformation on the \GN{} model \cite{Allamanis:Model:GGNN:CoRR:2017}, where the target task is to label the correct variable name based on the understanding of its usage.

Additionally, we have only experimented with a small set of examples \cite{Link:code2vec:Demo}. 
Our further plan includes a detailed study with a larger Java dataset \cite{Link:code2vec:Dataset} for the \CV{} model and a larger C\# dataset \cite{Link:GGNN:Dataset} for the \GN{} model.

\begin{comment}

Furthermore, we decide to plan an ablation study to observe the insight about which transformation can be more stressful for neural models. 
We also interested to see whether combining multiple transformations or applying transformations at certain place contributed most. 
\end{comment}

\section{Related Work}
\label{sec:related}
\begin{comment}
    Make a model to mistakenly change output is an active area of research. Several approaches have already been proposed such that test inputs synthesizing, program transformation, rules for semantically equivalent adversaries, etc. The goals of those approaches are to reveal the erroneous behaviors of neural models. This section highlights the literature of transformation-based testing related to our proposed approach. 
\end{comment}

%\subsection{Transformation-based Testing}

%The key fact behind the transformed test input is to verify the robustness of neural models and identify the unwanted behaviors of neural models.
Several approaches for transformation-based testing have been proposed, such as 
DeepTest \cite{Tian:Transformation:DeepTest:ICSE:2018} and 
COSET \cite{Wang:Transformation:COSET:arXiv:2019}.

Tian et al. \cite{Tian:Transformation:DeepTest:ICSE:2018} proposed DeepTest, a tool for automated generation of real-world test images and testing of DNN-driven autonomous cars.
They introduced potential image transformations (e.g., blurring, scaling, fog and rain effects) that mimic real-world conditions.
They applied transformation-based testing to identify the numerous corner cases that may lead to serious consequences, such as a collision in an autonomous car.
%The main difference between this work and ours is that their technique was for image transformation while we perform the transformation on programs where the semantic meaning of the transformed program must be correlated with the original program. 
Another study in this area was conducted by the authors of DeepRoad \cite{Zhang:Transformation:DeepRoad:ASE:2018}, who applied extreme realistic image-to-image transformations (e.g., heavy snow or hard rain) using the DNN-based UNIT method. 

Wang et al. \cite{Wang:Transformation:COSET:arXiv:2019} proposed COSET, a framework for standardizing the evaluation of neural program embeddings. They applied transformation-based testing to measure the stability of neural models and identify the root cause of misclassifications. 
They also implemented and evaluated a new neural model called LIGER \cite{Wang:Model:LIGER:arXiv:2019} with COSET's transformations, where they embedded programs with runtime information rather than learning from the source code. 

\bibliography{references}
\bibliographystyle{IEEEtranS}

\end{document}